\documentclass[journal]{IEEEtran}
\usepackage[utf8]{inputenc}
\usepackage{graphicx}
\usepackage[many]{tcolorbox}
\usepackage{natbib}
\usepackage{balance}
\bibpunct{(}{)}{,}{a}{}{;}
\begin{document}

\title{Textual Stylistic Variation: Choices, Genres and Individuals}
\author{Jussi Karlgren \\ Swedish Institute of Computer Science (SICS)}

\maketitle

\begin{tcolorbox}[colback=red!10!white,
                     colframe=red!20!black,
                     title=\textsc{This paper was originally published as a chapter in},  
                     center, 
                     valign=top, 
                     halign=left,
                     before skip=0.8cm, 
                     after skip=1.2cm,
                     center title, 
                     width=3in]
                      \textsc{The structure of style: Algorithmic approaches to understanding manner and meaning}, \\ edited by Shlomo Argamon, Kevin Burns, and Shlomo Dubnov. Springer. 2010.

  \end{tcolorbox}

\begin{abstract}
This chapter of the collection argues for more informed target metrics for the
statistical processing of stylistic variation in text
collections. Much as operationalized relevance proved a useful goal to
strive for in information retrieval, research in textual stylistics,
whether application oriented or philologically inclined, needs goals
formulated in terms of pertinence, relevance, and utility---notions
that agree with reader experience of text.  Differences readers are
aware of are mostly based on utility---not on textual
characteristics per se.  Mostly, readers report stylistic differences
in terms of genres.  Genres, while vague and undefined, are
well-established and talked about: very early on, readers learn to
distinguish genres. This chapter discusses variation given by genre,
and contrasts it to variation occasioned by individual choice.
\end{abstract}

\section{Stylistic variation in text}\index{text} \index{reading} \index{author choice}

Texts are much more than what they are about. Authors make choices
when they write a text: they decide how to organize the material they
have planned to introduce; they select amongst available synonyms and
syntactic constructions; they target an intended audience for the
text. Authors will make their choices in various ways and for various
reasons: based on personal preferences, on their view of the reader,
and on what they know and like about other similar texts.  These
choices are observable to the reader in the form of stylistic
variation, as the difference between two ways of saying the same
thing.

On a surface level this variation is quite obvious, as the choice
between items in a vocabulary, between types of syntactical
constructions, between the various ways a text can be woven from the
material it is made of. A consistent and distinguishable tendency to
make some of these linguistic choices can be called a \emph{style}.
It is the information carried in a text when compared to other texts,
or in a sense compared to language as a whole. It is not incidental
but an integral part of the intended and understood communication, and
will impart to the reader a predisposition to understand the meaning
of text in certain ways. Managing stylistic choice is the mark of a
competent author---learning how to do so well is a craft which
requires training and experience. Recognising stylistic choice is an
important component of reading competence---a proficient reader will
be sensitive to stylistic variation in texts.

Consistent situationally motivated strategies for making
the appropriate stylistic choices is a functional strategy on the part
of the author, and enables readers to identify likenesses across
individual texts or utterances, thus guiding the reader in
understanding them. Such consistent bundled observable occurrence patterns
of linguistic items 
are easily observable and recognisable by statistical analysis and constitute a useful tool for achieving the communicative purposes of the author. 

\section{Detecting stylistic variation in text} \index{textual variation}

It is easy enough to establish that there are observable stylistic
differences between texts we find in document collections. However,
while statistical analysis of stylistic differences between texts is
mostly uncomplicated in every particular instance, it is difficult to
provide general solutions, without descending into specifics or
idiosyncracies.  Using the textual, lexical, and other linguistic
features we find to cluster the collection---without anchoring
information in usage---we risk finding statistically stable
categories of data without explanatory power or utility.

A more linguistically informed approach is to start from established
knowledge (or established presumption, as it were) and to work with a
priori hypotheses on qualities of textual variation. The observations
we are able to make from inspection of the linguistic signal are
limited by the quality of the analysis tools we have recourse to. The
features we find are often on a low level of abstraction and are
simultaneously superficial and specific. To abstract to a higher level
of abstraction from the observable surface features it is useful to
understand them as the exponents of some underlying or latent
dimensions of variation within the space of possible texts---Douglas
Biber, for example, whose work has informed much of the work on
computational stylistics, has posited dimensions such as Involved vs
Informed, Narration vs Argumentation, Personal vs
Impersonal~\citep{Biber:88, Biber:89}.

Investigating such underlying dimensions of variation, establishing
hypotheses about which dimensions are reasonable and effective, based
on study of the material at hand, on knowledge of human linguistic and
communicative behaviour, on understanding of computation and
processing constraints, is the main task for computational stylistics
today---and a task which only to some extent can be accomplished
using purely formal, data-oriented methods, without studying either
textuality or readership.

If we wish to contribute to better text handling tools and to a better
understanding of human textual practice by the statistical processing
of stylistic variation in text collections we need principled
hypotheses of human linguistic processing and communicative behaviour
to base our experimentation on.  We need to understand what
consequences our claims have on representation, processing, and
application, and we need a research methodology which invites the
systematic evolution of new results. Much as operationalized relevance
has proven a useful goal to strive for in information retrieval,
research in textual stylistics, whether application oriented or
philologically inclined, needs goals formulated in terms of
pertinence, relevance, and usefulness.

Such notions agree with reader experience of text: the differences
readers are aware of are mostly based on utility---not on textual
characteristics per se. 

\section{Genres as vehicles for understanding stylistic variation} \index{textual genre} \index{genre}

Stylistic differences vary by author preferences and by constraints
imposed by situation the text is produced in. This is reflected when
readers are asked about texts.

In questionnaires or interviews, readers mostly report stylistic
differences either by describing the \emph{source} of the text, or in
terms of categories of text, in terms of \emph{genres}. Readers
describe genres in terms \emph{utility}, in terms of perceived
\emph{quality} of the text in some prescriptive terms, or in terms of
\emph{complexity} of the text and subject matter. On follow-up
questioning readers will bring up subjective qualities such as
\emph{trustworthiness} of the text or further discuss
\emph{readability} or other specifics related to complexity, both
lexical and syntactic (and specifically making claims in both
dimensions simultaneously, typically claiming that leader articles are
difficult and long-winded or that sports features are inane and
simplistic).  Readers will describe genres by describing situations in
which the genre is relevant, through experiences of reading a genre,
by typical topics or content, and only in some very specific cases do
readers give examples of lexical or other linguistic features of the
text~\citep{DeweEA:98}. 

Demarcation of stylistic variation to topical variation is of course
impossible: the content and form of the message cannot be
divorced. Certain meanings must or tend always to be expressed in
certain styles: legal matters tend to be written in legal jargon
rather than hexameter; car ownership statistics in journalistic or
factual style. Drawing a clean distinction between meaning and style,
between form and content, is in practice impossible except for certain
special cases; it is worth questioning whether there are any formally
identifiable styles at all beyond the distinctions that topical
analysis already give us \citep{Enkvist:73}.

Genre is a vague but well-established notion, and genres are
explicitly identified and discussed by language users.  Early on in
their reading career, readers learn to distinguish texts of different
genres from each other: children's books from encyclopedias, news from
magazines, handbooks from novels. Genres have a reality in their own
right, not only as containers or carriers of textual
characteristics. 

Genre analysis is a practical tool for the analysis of human activity
in many ways in many different fields loosely related to each other,
some more formal than other. In recent years, genre analysis has been
extended to typologies of communicative situations beyond the purely
textual \citep{Bakhtin:86, Swales:90}, with genres ranging much
further than would be afforded by analysis of written text: the range
of possible human communicative activities is much wider than the
range of possible texts. (This present discussion is mostly concerned
with written texts, but it should be noted that generalisations to
other communicative situations are obvious and obviously
interesting). Each such communicative sphere can be understood to
establish its conventions as to how communicative action can be
performed. When these conventions bundle and aggregate into a coherent
and consistent socially formed entity, they guide and constrain the
space of potential communicative expressions and form a genre,
providing defaults where choices are overwhelming and constraints no
choices should be given. 

A text cannot stray too far the prototypical expression expected
within the genre, or its readers will be confused and hindered in
their task of understanding it. Genre conventions form a framework of
recurrent practice for authors and readers alike and a basis of
experience within which the communication, its utterances and
expressions are understood: participants in the communicative
situation use style as a way of economizing their respective
communicative effort. This basic premise, that of effective
communicative practice on part of participants, should be kept in mind
during the study of stylistic variation in text: they will form be the
basis on which hypotheses of latent dimensions of appropriate
explanatory power can be built.

The perceptions of readers may be interesting, elucidating, and
entertaining to discuss, but are difficult to encode and put to
practical use. While we as readers are good at the task of
distinguishing genres, we have not been intellectually trained to do
so and we have a lack of meta-level understanding of how we proceed in
the task. To provide useful results, whether for application or
for the understanding of human communicative behaviour, research
efforts in stylistic analysis must model human processing at least on
a behavioural level at least to some extent. We need to be able to
address the data on levels of usefulness, and we need to observe
people using documents and textual information to understand what is
going on.

\section{Factors which determine stylistic variation in text} \index{stylistic variation in text} \index{linguistic convention} \index{conventionalisation} \index{grammaticalisation}

Stylistic choices are governed by a range of constraints, roughly
describable in terms of three levels of variation: firstly, and most
obviously observable, choices highly bound and formalized into
situation- and text-independent rule systems such as spelling,
morphology or syntax, taught in schools and explicitly mentioned in
lexica, grammars, and writing rules; secondly, choices that are
conventional and habitual, bound by context, situation, and genre,
mainly operating on the level of lexical choice and phraseology;
thirdly, choices on the level of textual and informational
organisation where the author of a text operates with the least amount
of formal guidance (cf. Table~\ref{ll}).

Constraints in the form of language rules are studied by linguists and are understood as obligatory by language users; linguistic conventions, bound by situation are more difficult to pinpoint and formulate, except as guidelines to an author in terms of appropriateness or even ``politeness''; constraints on informational organisation are few and seldom made explicit, but are what distinguishes a well-written text from a badly written one. 

\begin{table*}[t]
\caption{Characteristics of different levels of stylistic constraints.}
\label{ll}
  \centerline{
\begin{tabular}{|p{2cm}|p{3cm}|p{5cm}|}
\hline
\bf Rule & Language &  Syntax, morphology \\
\hline
\bf Convention &  Situation or genre &  Lexical patterns, argumentation structures, tropes \\
\hline
\bf Free    &       Author  &       Repetition, organisation,    elaboration \\
\hline
\end{tabular}
}
\end{table*}

The conventions that a genre carries are sometimes explicitly
formulated, consciously adhered to by authors, and expected by
readers. Other times they are implicit and not as clearly
enforced. They may even be misunderstood by the parties of a discourse
without causing a complete breakdown.

For the purposes of descriptive analysis of texts, genres can be
described through (a) the communicative purpose which should be
somewhat consistent within the set of texts that constitute it; (b) the
shared and generalised stylistic and topical character of the sets of
documents, which also should be somewhat consistent; (c) the contexts
in which the set of texts appear; (d) the shared understanding between
author and reader as to the existence of the genre.

This last aspect does not need to be explicit to all readers or even
all authors, but in typical cases, the genre is something that can be
discussed by experienced authors and experienced readers, and which is
intuitive and helpful to both parties. Items belonging to a genre are
recognized through their appearance in specific contexts and the
presence of some identifiable and characteristic choices. Phrases such
as ``Once upon a time'', ``In consideration thereof, each party by
execution hereof agrees to the following terms and conditions'';
rhyming or alliteration; expressions of personal judgment, opinion and
sentiment; explicit mention of dates and locations are all examples of
surface cues which are easy to identify. Presentation of texts printed
in book form, as hand written pamphlets, read aloud at a fireside or
at a lectern in a congress hall are contextual cues that afford us
licence to assume them to belong to families of other similar items.

Establishing a genre palette for further studies can be a research
effort in its own right. Understanding how genres are established and
how they evolve in the understanding participants have and acquire of a
communicative situation is a non-trivial challenge and requires
elaborate methodological effort. Examples of such studies are studies of communicative patterns in distributed organisations \citep{Yates:Orlikowski:92, Orlikowski:Yates:94} or studies of how genres can be used to understand communication on the web. \citep{webgenre}

However, understanding the social underpinnings of genres is not
necessary for most computational approaches. The genres used as target
metrics must have some level of salience for their readers and
authors, but they must not cover all possible facets and niceties of
communication to be useful for the fruitful study and effective
application of stylistic variation in text. In many experiments made
on computational stylistics, in the case of many experimental
collections, and in fact in the minds of many readers, genre has
mostly been equated with or based on text source. Since genres are
strongly dependent on context, and authorship analysis is one of the
prime applications for stylistic analysis, this is a fair
approximation. More usage-based and functionally effective approximations than source can be high level categories such as ``narrative'' vs ``expository''; ```opinionated'' vs ``neutral''; ``instructional and factual'' vs ``fiction and narrative'' vs ``poetry and drama''; or alternatively, specific categories such as ``legal'', ``technical'', ``commerc ial'' as might be found in some specific collection such as a corporate document archive. 
The central point is that the categories chosen must have some base in usage rather than in statistically determined differences: the latter can be overspecific, underdetermined, and lead the study astray.

\section{Individual variation vs situational variation} \index{indvidual style vs functional style}

Seasoned authors are likely to feel free to break many of the
conventions of a genre, in effect creating a voice or style of their
own (or even a style specific to some specific text or situation),
where a novice author will be more likely to follow conventions,
falling back on defaults where experience gives insufficient guidance,
using unmarked cases where choice between alternatives is
difficult. (Cf. to Burns discussing style as a guiding mechanism for poker players in Chapter 12 of this volume \citep{burns2010style}, and to Cohen discussing style in art in Chapter 1 of this volume \citep{cohen2010style}.) In this sense, genre gives us a benchmark or a water-line
with which to judge individual choice, measuring contrast between {\em
functional style}, which forms the identifying characteristics of the
genre, in contrast with {\em individual style} on the level of
specific texts or sources \citep{Vachek:75}.

The span of variation, from variation occasioned by individual
performance to that constrained by the expectations given by genre,
gives us several targets for stylistic study: we may wish to explore
the general characteristics of communicative situations of some
specific type, domain or topic; or we may wish to understand the character of a
specific author or individual source. After selecting a linguistic
item---some lexical item, some construction, some observation---
we can study if its occurrence pattern in some sample set of texts varies
from the expected occurrence of that specific item (with prior information
taken into account). This is a mark of individuality and
can be used in the process of identifying author, genre, or indeed, topic.

\section{Concrete Example: News Text and Its Subgenres} \index{style in newsprint}

News text, while a special register of human language use in itself,
is composed of several well-established subgenres. While news text is
---together with scientific and intellectual text---over-represented
as an object of empirical philological study as compared to other
forms of human linguistic communication, it possesses some quite
useful qualities for systematical study of linguistic characteristics,
chief among them that the texts are most often intended to be the way
they are. The textual material has passed through many hands on its
way to the reader in order to best conform to audience expectations;
individual variation---worth study in itself---is not preserved to
any appreciable extent in most cases; the audience is well trained in
the genre. In short, the path is well-trod and as such easy to study.
In this example one year of
Glasgow Herald is studied, and many of the texts are tagged for
article type---see Table~\ref{g}.

\begin{table}[bp]
\begin{center}
\begin{tabular}{|rrr|}
\hline
\bf Article type & & $n$ \\
\hline
{\bf tagged}    & &              17~467 \\ 
advertising     &               522 & \\ 
book    &               585 & \\ 
correspondence  &               3~659 & \\ 
feature         &               8~867 & \\ 
leader  &               681 & \\ 
obituary        &               420 & \\ 
profile         &               854 & \\ 
review  &               1~879 & \\ 
\hline 
{\bf untagged}    & &              39~005 \\ 
\hline 
{\bf total}   &    &           56~472 \\ 
\hline 
\end{tabular}
\end{center}
\caption{Sub-genres of the Glasgow Herald, with the number of articles in each.}\label{g}
\end{table}

It is easy enough to establish that there are observable differences
between the genres we find posited in the textual material. Trawling
the (morphologically normalized) texts for differences, comparing each
identified category in table*~\ref{g} with the other categories we find
for a one month sample of text some of the most typical words for each
category as per Table~\ref{o}. ``Typical'' is here operationalized as
words with a document frequency deviating from expected occurrence as
assessed by $\chi^{2}$. This sort of analysis can be entertaining, at
times revealing, but cannot really give us any great explanatory
power. Even the handful of example terms shown in Table~\ref{o} are
clearly coloured by the subject matter of the sub-genres and only to
some extent reveal any stylistic difference between them.

\begin{table}[bp]
\begin{center}
\begin{tabular}{|ll|}
\hline
\bf Article type & \bf Typical words \\
\hline
advertising     &  provide, available, service, \\
                & specialist, business\\ 
book review     &  novel, prose, author, \\ 
& literary, biography, write \\ 
correspondence  &  (Various locations in Scotland), \\ 
& overdue, SNP \\ 
feature         &  say, get, think, put, there, \\ 
& problem, tell \\ 
leader          &  evident, government, outcome,  \\
& opinion, even\\ 
obituary        &  church, wife, daughter, \\
& survive, former \\ 
profile         &  recall, career, experience, \\
& musician \\ 
review          &  concert, guitar, piece, \\
& beautifully, memorable \\ 
\hline 
\end{tabular}
\caption{Typical words in sub-genres of one month of the Glasgow Herald.}\label{o}
\end{center}
\end{table}

\section{Measurements and Observanda} \index{stylostatistics}

Measurement of linguistic variation of any more sophisticated kind has
hitherto been hampered by lack of useful tools. Even now, most
measurements made in text categorisation studies, be it for the purposes of authorship
attribution, topical clustering, readability analysis or the like,
compute observed frequencies of some lexical items, or some
identifiable constructions. These measurements are always
\emph{local}, inasmuch they only take simple appearances of some
linguistic item into account. Even when many such appearances are
combined into an average or an estimate of probability of recurrence,
the measurement in question is still local, in the sense that
they disregard the structure and progression of the communicative
signal. An observed divergence in a text sample
from the expected occurrence rate of that specific item (with prior
information taken into account) is a mark of specific character and
can be used in the process of identifying, e.g., topic, genre, or
author. 

The examples given above in Table~\ref{i} are of this kind. A
systematic tendency to pursue certain choices on the local level may
be symptomatic of an underlying dimension, but is hardly likely to
have any more far-reaching explanatory or predictive power.
It is difficult to systematically explore, find, and formulate
sensible or adequate dimensions on any level of abstraction above the
trivial if the features we have at our disposal are limited to average
occurrence frequency of pointwise observations of simple linguistic
items.

Textual variation might instead profit from being measured on levels on which authors
and readers process information, not on the level of whitespace
separated character strings. Tentative candidates for more textual
measurements could be using term recurrence rather than term frequency
\citep{Katz:96}; term patterns \citep{Sarkar}; type-token ratio
\citep{Tallentire}; rhetorical structure; measures of textual cohesion
and coherence; measures of lexical vagueness, inspecificity, and
discourse anchoring; and many other features with considerable
theoretical promise but rather daunting computational requirements.

Leaving the simple lexical occurrence statistics and reaching for
features with better informational potential proves to be rather
unproblematic. We know from past studies and judicious introspection
that interview articles contain quotes and that leader articles
contain argumentation; both are to be expected to contain pronouns and
overt expressions of opinion. Table~\ref{i} shows some such
measurements\footnote{The `{\sc private}' type of factual verb expresses intellectual states such as belief and intellectual acts such as discovery. These states and acts are `private' in the sense that they are not observable... Examples of such verbs are: {\it accept, anticipate ... fear, feel, ... think, understand} \citep{QuirkEA:85}. Opinion expression is counted by adverbials such as \emph{clearly, offend, specious, poor, excellent.} and argument expression by counting argumentative operators such as \emph{if ... then, else, almost.}}. The explanatory power of this
table---again, while interesting in some respects and worth
discussion---is rather low, and application of the findings for
prediction of genre for unseen cases will be unreliable. A more
abstract level of representation is necessary to be able to extract
useful predictions and to gain understanding of what textual variation
is about.

\begin{table*}[tbp]
\centerline{
{\small
\begin{tabular}{|r||c|c|c|c|c|}
\hline
 & Personal  & Demonstratives & Utterance \&  & Opinion & Characters  \\
 & pronouns  & ``that'' \&c & private verbs & \& argument & per word  \\
\hline
advertising     & -     & .     & +      & .  & + \\
book review     & +     & +     & +      & +  & - \\
correspondence  & -     & -     & -      & -  & + \\
feature         & +     & +     & +      & +  & - \\
leader          & -     & +     & .      & +  & + \\
obituary        & -     & -     & .      & -  & - \\
profile         & +     & +     & +      & +  & - \\
review          & -     & -     & -      & -  & + \\
\hline 
\end{tabular}
}
}
\vskip1em
{\footnotesize
\begin{tabular}{rl}
\bf Key: & \\
+ &  Significantly higher values  \\
- &  Significantly lower values  \\
. & Non-significant value distribution \\
\end{tabular}
}
\caption{Some measurements of linguistic items per sub-genre of one year of the Glasgow Herald. Statistical significance ($p > 0.95$) by Mann-Whitney U.}\label{i}
\end{table*}

\section{Aggregation of Measurements}

A text will typically yield a number of observations of a linguistic
item of interest. Most experiments average the frequency of
observations and normalise over a collection of texts, or calculate
odds and compare to estimated likelihoods of occurrence, as the
example given above. Aggregating observations of linguistic items by
averaging local occurrence frequencies over an entire text does not utilise the specific character of text, namely that it is a non-arbitrary sequence of symbols. 

There is no reason to limit oneself to pointwise aggregation of
observations: the inconvenience of moving to other aggregation models
is minimal. In another experiment using the same data as above we have
shown that \emph{text configurational} aggregation of a feature yields
better discriminatory power than pointwise aggregation does
\citep{KarlgrenEriksson:07}. 

To obtain simple longitudinal patterns each observed item is 
measured over sliding windows of one to five sentences along each
text, and the occurrence of the feature is recorded as a {\em
  transition pattern} of binary occurrences, marking the feature's
absence or presence in the sentences within the window. The first and
last bits of text where the window length would have extended over the
text boundary can be discarded. The feature space, the possible values
of the feature with a certain window size consists of all possible
transition patterns for that window size---yielding a feature space of size $2^n$ for a window size of $n$. For windows of size two, the
feature space consists of four possible patterns, for windows of size
five, thirty-two. 

\section{Concrete example: Configurational features} \index{configurational features}

Our hypothesis is that author (and speaker) choice on the level of
informational structuring and organisation (see, again,
Table~\ref{ll}) is less subject to pressure from conventionalisation
and grammaticalisation processes\footnote{This experiment is reported
in full in \citep{KarlgrenEriksson:07}. This section is an excerpt of
that paper.}. This is both by virtue of wide scope, which limits the
possibilities of observers to track usage, as well as by the many degrees
of freedom open for choice, which makes rule expression and rule
following inconvenient.

We believe that configurational aggregation of observations might
improve the potential for categorisation of authors, since they
preserve some of the sequential information which is less constrained
by convention and rules. The experiment is designed to investigate
whether using such longitudinal patterns improves the potential for
author identification {\em more} than it improves the potential for
genre identification: these transition patterns can then be compared
for varying window lengths---the operational hypothesis being that a
longer window length would better model variation over author rather
than over genre.

Using the same data from Glasgow Herald, we calculated a simple binary
feature, noting the occurrence of more than two clauses of any type in
a sentence.  Each sentence was thus given the score 1 or 0, depending
on whether it had several clauses or only one. This feature is a
somewhat more sophisticated proxy for syntactic complexity than the
commonly used sentence length measure.

The measurements are given in Table~\ref{t1} for all genres, and the
authors with the highest and lowest scores for each variable. As seen
from the table, genres are more consistent with each other than are
authors: the range of variation between 0.52 to 0.96 is greater than
that between 0.78 and 0.93.

\begin{table*}[h]
\begin{center}  
\begin{tabular}{|r|c|c|}
\hline
category & single-clause ($f_0$)  & multi-clause ($f_1$) \\    
\hline                                 
advertising      & 0.90 & 0.10 \\   
book             & 0.83 & 0.17 \\ 
correspondence   & 0.92 & 0.08 \\ 
feature          & 0.93 & 0.07 \\ 
leader           & 0.93 & 0.07 \\ 
obituary         & 0.78 & 0.22 \\ 
profile          & 0.92 & 0.08 \\ 
review           & 0.87 & 0.13 \\ 
\hline                                 
author $A_M$   & 0.96 & 0.04  \\
author  $A_m$   & 0.52  & 0.48 \\
\hline 
\end{tabular}
\end{center}
\caption{Relative presence of multi-clause feature ``clause'' in sentences.}\label{t1}
\end{table*}

\begin{table*}[h]
\begin{center}
\begin{tabular}{|l|llll|}
\hline
genre & $f_{11}$ & $f_{10}$ &  $f_{01}$ &  $f_{00}$ \\
\hline
advertising     & 0.011         & 0.072         & 0.039         & 0.88 \\
book            & 0.038         & 0.084         & 0.069         & 0.81 \\
correspondence  & 0.066         & 0.15          & 0.051         & 0.73 \\
feature         & 0.022         & 0.078         & 0.056         & 0.84 \\
leader          & 0.016         & 0.055         & 0.023         & 0.91 \\
obituary        & 0.0079        & 0.072         & 0.023         & 0.90 \\
profile         & 0.016         & 0.056         & 0.040         & 0.89 \\
review          & 0.041         & 0.13          & 0.072         & 0.76 \\
\hline 
author & $f_{11}$ & $f_{10}$ &  $f_{01}$ &  $f_{00}$ \\
\hline
$A_1$ &  0.013   &        0.071 &       0.052    & 0.86 \\      
$A_2$ &  0.021   &        0.050 &       0.018    & 0.92 \\      
$A_3$ &  0.018   &        0.11  &       0.088    & 0.78 \\      
$A_4$ &  0.19    &        0.097 &       0.032    & 0.68 \\      
$A_5$ &  0.013   &        0.11  &       0.052    & 0.82 \\      
$A_6$ &  0.0062  &        0.071 &       0.020    & 0.90 \\      
$A_7$ &  0.018   &        0.063 &       0.038    & 0.88 \\      
$A_8$ &  0.0067  &        0.047 &       0.032    & 0.91 \\      
$A_9$ &  0.010   &        0.064 &       0.027    & 0.90 \\ 
\hline 
\end{tabular}
\end{center}
\caption{Relative frequency of multi-clause sentence sequences for genres and some authors, window size 2.}\label{g2}
\end{table*}

Expanding the feature space four-fold the relative presence for the
various sequences of multi-clause sentences, in a window of size two,
are shown in Table~\ref{g2} for the genres labeled in the data set and
for some of the more prolific authors. 

This table cannot be directly compared to the measurements in
Table~\ref{t1}, but using the frequency counts as probability
estimates, we can use Kullback-Leibler divergence measure~\citep{KL} as
a measure of differences between measurements. The Kullback-Leibler
divergence is a measure of distance between the states in the
probability distribution, a large divergence indicating better
separation between states--which is desirable from the perspective
of a categorisation task, since that would indicate better potential
power for working as a discriminating measure between the categories
under consideration. Since the measure as defined by Kullback and
Leibler is asymmetric, we here use a symmetrised version, a harmonic
mean~\citep{KLS-Symm}. In this experiment, with eight genre
labels and several hundred authors, we perform repeated resampling of
eight representatives, fifty times, from the set of authors and
average results to obtain comparable results with the genre
distribution.

For each window length, the sum of the symmetrised Kull\-back-Leibler
measure for all genres or authors is shown in Figure~\ref{kls}. The
figures can only be compared horizontally in the table---the
divergence figures for different window sizes (the rows of the table),
cannot directly be related to each other, since the feature spaces are
of different size. This means that we cannot directly say if window
size improves the resulting representation or not, in spite of the
larger divergence values for larger window size. We can, however, say that
the {\em difference} between
genre categories and author categories is greater for large window sizes.
This speaks to the possibility of our main hypothesis holding: a larger window size
allows a better model of individual choice than a shorter one.

\begin{table}[t]
  \centerline{
\begin{tabular}{|r|rr|}
\hline
Window size     &  Genre        &        Author \\
\hline
1       & 0.5129 & 0.7254 \\
2       & 0.8061 & 1.3288 \\
3       & 1.1600 & 2.1577 \\
4       & 1.4556 & 2.3413 \\
5       & 1.7051 & 3.0028 \\
\hline
\end{tabular}
}
\caption{Window size effect.}\label{kls}
\end{table}

\balance

\section{Conclusion: Target Measures}

Our example experiment shows how the choice of observed linguistic
item, choice of aggregation method, and choice of target hypothesis
all work together towards making sustainable statements about
stylistic variation.

The experiment shows that one method of aggregation gives different
results from another set; they also show that modelling author
behaviour can profitably be modelled by different features than genre---presumably, in the one case, identifying conventions, in the other,
avoiding them.

In general, only if useful target measures of pertinence, relevance,
and utility can be found for evaluation, couched in terms derived from
study of readership and reader assessment of texts for real needs can
we hope to accomplish anything of lasting value in terms of
understanding choice, textual variation, and reading. This, in turn,
can only be achieved beginning from an understanding that stylistic
variation is real on a high level of abstraction, in the sense that
readers, writers, editors all are aware of genres as an abstract
category; that genres are a useful mid-level category to fix an
analysis upon, and that there is no independent non-arbitrary genre
palette outside the communicative situation the study is modelling.

Results of any generality can only be achieved with more informed
measures measures of variation and choice, couched in terms of
linguistic analysis rather than processing convenience, and aggregated
by computational models related to the informational processes of the
author and reader.

\bibliographystyle{authordate1}
\bibliography{Karlgren_Structure_of_Style}

\end{document}